%% file: Paper.tex
\documentclass{article}
\usepackage{spconf,amsmath,graphicx,amssymb}

\usepackage{epstopdf}
\usepackage{caption}
\usepackage{subcaption}
\usepackage{array}

\newcolumntype{L}[1]{>{\centering\let\newline\\\arraybackslash\hspace{0pt}}m{#1}}

\title{A Deep Learning Approach to Multiple Kernel Fusion}
\makeatletter
\def\@name{ \emph{Huan Song$^{\dagger}$, Jayaraman J. Thiagarajan$^{\ddagger}$, Prasanna Sattigeri$^{\star}$,}\\
 \emph{Karthikeyan Natesan Ramamurthy$^{\star}$ and Andreas Spanias$^{\dagger}$}
 \thanks{This research was supported in part by the SenSIP center.}}
\makeatother
\address{$^{\dagger}$ SenSIP Center, ECEE, Arizona State University, Tempe, AZ \\
$^{ \ddagger}$ Lawrence Livermore National Labs, 7000 East Avenue, Livermore, CA \\
$^{\star}$ IBM T.J. Watson Research Center, 1101 Kitchawan Road, Yorktown Heights, NY}
\begin{document}
%
\maketitle
\begin{abstract}
Kernel fusion is a popular and effective approach for combining multiple features that characterize different aspects of data. Traditional approaches for Multiple Kernel Learning (MKL) attempt to learn the parameters for combining the kernels through sophisticated optimization procedures. In this paper, we propose an alternative approach that creates dense embeddings for data using the kernel similarities and adopts a deep neural network architecture for fusing the embeddings. In order to improve the effectiveness of this network, we introduce the kernel dropout regularization strategy coupled with the use of an expanded set of composition kernels. Experiment results on a real-world activity recognition dataset show that the proposed architecture is effective in fusing kernels and achieves state-of-the-art performance.
\end{abstract}
\section{Introduction}
\input{intro}
\section{Proposed Approach}
\input{approach}
\section{System Setup}
\input{setup}
\section{Performance Evaluation and Conclusion}
\input{evaluation}

\clearpage

\bibliographystyle{IEEEbib}
\bibliography{refs}

\end{document}

%% file: intro.tex
Kernel methods provide a powerful framework to extend several machine learning formulations since they enable the design of effective non-linear models. For example in Support Vector Machines (SVM), the problem of building binary classifiers to obtain non-linear decision boundaries can be reposed into a dual problem in terms of the kernel similarity matrix. Referred to as the kernel trick, this approach has been successfully applied to a wide range of supervised and unsupervised learning problems \cite{yan2005graph,elisseeff2001kernel}. A valid positive semidefinite kernel inherently defines a lifting (transformation) to a Reproducing Kernel Hilbert Space (RKHS), thereby enabling efficient approximation of any function of interest in the transformed space. Another important property of kernel methods is that fusing kernels from multiple sources (e.g. different feature descriptors or sensing modalities) is straightforward. A commonly adopted strategy is to consider a convex combination of the kernels. The process of simultaneously inferring the weights for the convex combination and minimizing the structural risk (SVM objective) is referred to as Multiple Kernel Learning (MKL) \cite{jain2012spf,sun2010multiple}. The idea is to effectively exploit the complementary nature of the different features and the representation power of different kernel functions. Despite their wide-spread use, as pointed out by \cite{gonen2008localized}, MKL algorithms may suffer when solving for global weights and the most critical support vectors, since the weight for a kernel is restricted to be the same over the whole input space. This challenge is alleviated using Localized MKL (LMKL) \cite{gonen2008localized}, which introduces a gating function for each kernel. By treating the input data sample as a variable, the gating function is able to characterize the underlying localities in data and promotes reduced number of support vectors.


\begin{figure}[t]
  \centering
  \centerline{\includegraphics[scale=0.63]{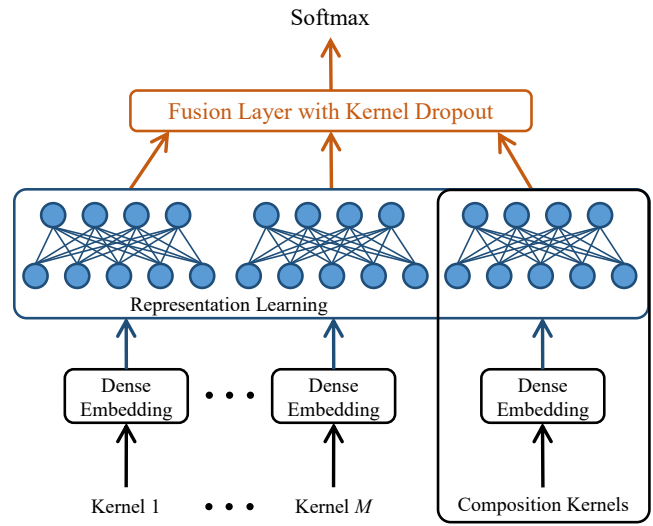}}
%
\caption{Proposed approach for multiple kernel fusion. We use the kernel matrices to create dense embeddings for data and fuse them in a fully connected deep network. The kernel dropout regularization at the fusion layer and the use of different composition (sum) kernels improves the optimization.}
\label{fig:framework}
\end{figure}

Several existing approaches for feature fusion begin by building compact and effective representations from raw features since fusing such compact representations can be robust to noise and outliers. In particular, sophisticated representation learning paradigms such as deep learning have shown exceptional power when dealing with complex, high-dimensional data. In \cite{ngiam2011multimodal}, the authors focused on feature learning for different multimodal settings and showed that in the multimodal fusion case, the fused feature exploits complementary information from each modality. In \cite{zhao2015heterogeneous}, Zhao \textit{et.al.} built sub-networks for each heterogeneous feature and relied on the Stacked Denoising Autoencoders to learn high-level homogeneous representations for feature integration. Note that, both methods start with the raw features directly and did not exploit the expressive power of similarity kernels. Existing works on incorporating the advantages of deep learning into kernel methods either develop novel kernel constructions to mimic the large neural computation \cite{cho2009kernel} or apply similar neural network structure to combine kernels and optimize at each layer \cite{strobl2013deep}.

In this paper, we propose to exploit the advantages of deep architectures in feature learning to build a new approach to multiple kernel fusion. First, we adopt a novel viewpoint to kernels by treating the similarities encoded in the kernel matrix as a valid embedding of the data. This is similar to the approaches in the natural language processing literature, wherein relevance measures such as Pointwise Mutual Information (PMI) of a word with respect to other words in the vocabulary is treated as a word embedding \cite{levy2014neural}. Since the kernel matrix can be inherently sparse and low-rank, we propose to apply an additional dense embedding layer (e.g. Singular Value Decomposition) to the columns of the kernel matrix. Consequently, the problem of kernel fusion is transformed to fusing their dense embeddings. To this end, we build a deep architecture for kernel fusion, coupled with novel training strategies: (a) to emulate the convex combination approach in MKL, we expand the set of input kernels by considering combinations of different subsets of the base kernels, and (b) we perform kernel dropout in the fusion layer for improved regularization. The proposed architecture replaces the complex optimization procedure in MKL by efficient representation learning and straightforward feature merging. This makes our fusion approach easily scalable to a large number of kernels.

%% file: approach.tex
\subsection{Architecture}
The proposed approach considers the similarity information encoded in a kernel as an embedding of the data, and poses the problem of MKL as fusing these embeddings in a deep learning architecture. In this section, we start by presenting the general architecture and then describe strategies for improving the performance.

As shown in Figure \ref{fig:framework}, the proposed architecture consists of three components: (a) obtain dense embeddings for data using kernel similarities, (b) representation learning using a deep architecture, and (c) feature fusion. Let us denote the kernel Gram matrix as $\mathbf{K}$, where $\mathbf{K}_{i,j}=k(\mathbf{x}_i,\mathbf{x}_j)$. Each column $\mathbf{s}_j$ of the Gram matrix encodes the relevance between sample $\mathbf{x}_j$ and all other samples $\mathbf{x}_i$ and it can be treated as an embedding for $\mathbf{x}_j$. This viewpoint is very similar to the construction of dense word embeddings using the PMI in text processing \cite{bouma2009normalized}. In the ideal case, $\mathbf{s}_j$ has large values for the samples that come from the same class with $\mathbf{x}_j$ and zeros for other samples. The sparsity in these embeddings makes them unsuitable for inference tasks \cite{levy2014neural}. To alleviate this, we obtain a dense embedding of the kernel similarities using Principle Component Analysis (PCA), which projects the original kernel feature to a low-dimensional space. Note that, this can be easily replaced by other dense embedding techniques including manifold learning \cite{elgammal2004inferring}, Word2Vec \cite{mikolov2013distributed} or random projection \cite{bingham2001random}. Besides providing dense embeddings, this step also helps to significantly improve the network training speed.  

On top of each dense feature set obtained by PCA, we build a fully connected neural network. The goal is to use back-propogation in a large network to learn a concise representation which will be more effective for inference tasks. To achieve this, the size of the network needs to be adequately large. In our application, we build a $4-$layer network separately for each embedding. At each hidden layer, dropout regularization \cite{srivastava2014dropout} is used to prevent overfitting and batch normalization \cite{ioffe2015batch} to accelerate training. After the representations are learned, we stack another layer which is responsible for fusing the features and obtaining the classification result with a softmax activation. The most straightforward approach for the feature merging is to simply concatenate all the inputs to the layer. However various other merge modes can be easily applied too including summation, averaging, multiplication etc. The flexibility of the merge layer facilitates a wide range of kernel combination forms. 

\subsection{Using Composition Kernels}
An important property of MKL is the various parameterization forms for mixing kernels such as convex combination, Hadamard product or mixtures of polynomials \cite{jain2012spf}. We emulate this property by including all possible combinations of base kernels (namely the composition kernels in Figure \ref{fig:framework}) to the architecture input. Given the base kernels set $\{ \mathbf{K}_1,\mathbf{K}_2,...,\mathbf{K}_M \}$, the whole input kernel set $\Phi$ to our architecture will have size $\tilde{M}$:
\begin{gather*}
\label{kerSet}
\Phi=\{\mathbf{K} \mid \mathbf{K}=\sum_{i=p}^{q}\mathbf{K}_i,\forall p,q\in \{1,2,3,...,M\},q\geq p \} \\
\tilde{M}=\sum_{m=1}^M\binom{M}{m}
\end{gather*}
Simple kernel summation proves to be highly effective in practical recognition problems and the derived representation corresponding to the summed kernel is often very different from either alone. Note that other formulations with base kernels can also be used. Paired with the flexible merge and deep feature representation, our architecture covers a large number of kernel combination scenarios without explicitly formulating them. 

\subsection{Kernel Dropout Regularization}
In dropout regularization \cite{srivastava2014dropout} for training large neural networks, neurons are randomly chosen to be removed from the network along with their incoming and outgoing connections. The process can be viewed as sampling a large set of possible network architectures with shared weights. Given the large kernel set $\Phi$, a more effective regularization mechanism is needed to prevent the network training from overfitting certain kernels. More specifically, we propose to regularize the fusion layer by dropping the entire representations learned from some randomly chosen kernels. Denoting the representations learned for all kernels as $\tilde{\Phi}=\{\mathbf{r}_1,\mathbf{r}_2,...,\mathbf{r}_{\tilde{M}}\}$ and a vector $\mathbf{t}$ associated with $\tilde{M}$ independent Bernoulli trials, the representation $\mathbf{r}_m$ is dropped from the fusion layer if $t_m$ is 0. The feed-forward operation can be expressed as:
\begin{gather*}
t_m\sim \text{Bernoulli}(p) \\
\hat{\Phi}=\{\mathbf{r}\mid \mathbf{r} \in \tilde{\Phi} \text{ and } t_m>0\} \\
\tilde{\mathbf{r}}=(\mathbf{r}_i),\mathbf{r}_i\in \hat{\Phi} \\
z_i=\mathbf{w}_i\tilde{\mathbf{r}}+b_i, y_i=f(z_i)
\end{gather*}
where $\mathbf{w}_i$ are the weights for hidden unit $i$, $(\cdot)$ denotes vector concatenation and $f$ is the softmax activation function. 

%% file: setup.tex
In this section, we apply the proposed architecture to the important problem of sensor-based activity recognition. Recent advances in activity recognition have shown promising results in the applications of fitness monitoring and assistive living \cite{zhang2013human}. However, problem still exists on how to effectively deal with the measurement inaccuracy and noise. One popular approach to the problem is utilizing various features and kernels that characterize salient aspects of the data and develop efficient fusion mechanisms to combine them. In this paper, we construct kernels which describe the statistical property, periodic structure and inter-sample relations for the accelerometer signals.

\subsection{Feature Extraction and Kernel Construction}
\label{kerCons}
\subsubsection{Statistics Kernel}
Statistical features have been known to be useful for activity recognition \cite{zhang2013human}. The features we use include mean, median, standard deviation, kurtosis, skewness and total acceleration. In addition, we extract the mean-crossing rate and dominant frequency to capture the frequency-domain information. We construct a Gaussian kernel and the best $\gamma$ parameter is determined by cross validation on the training set. 

\subsubsection{Shape Kernel}

In lieu of building conventional state-space models, Time Delay Embeddings (TDE) provide an effective way to reconstruct the underlying dynamical system from the observed data. Given a time-series data, the phase space is the set of states which contain all the necessary information to predict the future of the system  \cite{frank2010activity}. The TDEs of a time-series data $\mathbf{x}$ can be defined in matrix form $\mathbf{O}$ whose $i$th column is $\mathbf{o}_t=[x_t,x_{t+\tau},x_{t+2\tau},...,x_{t+(n-1)\tau}]$.


The $n$ time-delayed observation samples can be considered as points in $\mathbb{R}^n$, which is referred to as the delay embedding space. In our application, the delay parameter $\tau$ is fixed to $10$ and embedding dimension $n$ to $8$. Following the approach in \cite{frank2010activity}, we use PCA to project the embedding to 3-D for noise reduction. We extract a simple shape function based on the geometric distances, and use it to derive our feature. The shape function we consider measures the pair-wise distances between samples in the TDE space, calculated as $\mathbf{S}_{ij}=\|\mathbf{o}_i-\mathbf{o}_j\|_2$ \cite{venkataraman2016shape}. A histogram is calculated using these distances with a pre-specified bin size to build the feature. Following this, we construct an intersection kernel $k(\mathbf{h},
\mathbf{h^\prime})=\sum_i\min(h_i,h_i^\prime)$ \cite{maji2008classification}, where $\mathbf{h}$,$\mathbf{h^\prime}$ are the computed histograms.



\subsubsection{Correlation Kernel}

Correlation measures the dependence between two time-series signals and has been widely used in electroencephalogram (EEG) signal analysis. We calculate the absolute value of Pearson correlation coefficient. To account for shift between the two signals, the maximum absolute coefficient for a range of shift values is identified. The correlation matrix $\mathbf{R}$ defined in this way does not guarantee the required positive semi-definite condition of kernel. To correct this, we remove the negative eigenvalues from the matrix. Given the eigen-decomposition of the correlation matrix $\mathbf{R}=\mathbf{Q}\mathbf{\Lambda}\mathbf{Q}^{T}$, where $\mathbf{\Lambda}=\text{diag}(\lambda_1,\lambda_2,...,\lambda_n)$ and $\lambda_1\geq\lambda_2\geq...\geq\lambda_r\geq0\geq\lambda_{r+1}\geq...\geq\lambda_n$, the correlation kernel is constructed as $\mathbf{K}=\mathbf{Q}\mathbf{\hat{\Lambda}}\mathbf{Q}^{T}$, where $\mathbf{\hat{\Lambda}}=\text{diag}(\lambda_1,\lambda_2,...,\lambda_r,0,...,0)$.

\subsection{Dataset}
The dataset used in our experiments is obtained from \cite{zhang2012usc} and corresponds to $12$ different daily activities for $14$ subjects. Each activity is repeated in $5$ trials for each subject. The $3$-axis accelerometer measurements were obtained at a sampling rate of $100$Hz. We consider $5$ seconds of non-overlapping frames and as a result there are $5353$ frames. In our experiment, $80\%$ randomly chosen samples were used for training and the rest for testing. Putting together the $3$ base kernels described in Section \ref{kerCons} with the combination kernels ($\mathbf{K}_1+\mathbf{K}_2$, $\mathbf{K}_1+\mathbf{K}_3$ etc.) makes the total number of input kernels for our architecture to $7$.

%% file: evaluation.tex
The proposed approach is tested on the described activity recognition dataset. In the dense embedding stage, the dimension of the kernel feature is reduced to $500$. The $4$-layer neural network for representation learning has size $256$-$1024$-$512$-$64$. At each hidden layer the dropout rate is fixed at $0.5$. In the fusion layer, the kernel dropout rate is set to $0.6$. We use \texttt{Keras} library with \texttt{TensorFlow} backend to build and perform optimization for the architecture.

\begin{figure}[t]
  \centering
  \centerline{\includegraphics[width=7cm]{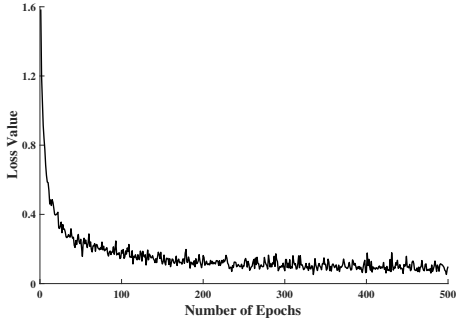}}
%
\caption{Convergence behavior of the proposed method.}
\label{fig:convergence}
\end{figure}

\begin{figure}[t]
  \centering
  \centerline{\includegraphics[width=6cm]{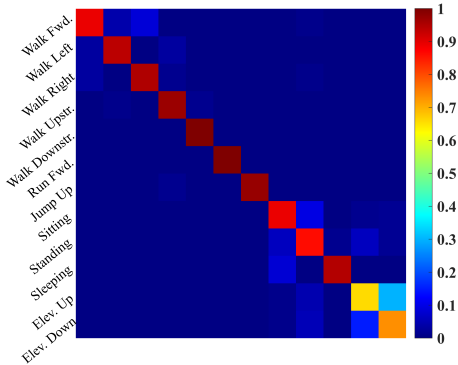}}
%
\caption{Confusion matrix based on the classification result.}
\label{fig:confusionMat}
\end{figure}

The training convergence curve shown in Figure \ref{fig:convergence} demonstrates that the architecture is able to reduce the loss value and achieve convergence quickly. We report the classification performance in Table \ref{table:restable}. In our case, the accuracy is defined as the averaged fraction of correctly predicted labels for all classes. We make comparison of our proposed architecture to other $3$ setups: (1) single kernel performance, which is obtained without the fusion layer, (2) different deep architecture settings, and (3) existing MKL methods. 

\begin{table}[t]
\caption{Classification Performance Comparison}
\centering
\begin{tabular}{|L{5cm}|L{2.5cm}|}
\hline
\textbf{Input Kernel} & \textbf{Accuracy ($\%$)} \\
\hline
\hline
Statistics &  79.3 \\
\hline
Shape & 73.3 \\
\hline
Correlation & 75.3 \\
\hline
\hline
\textbf{Architecture} & \textbf{Accuracy ($\%$)} \\
\hline
\hline
(a): Standard Feature Fusion & 82.3\\
\hline
(b): (a) + Dense Embedding & 86.6 \\
\hline
(c): (b) + Composition Kernels & 88.1\\
\hline
\hline
\textbf{MKL Method} & \textbf{Accuracy ($\%$)} \\
\hline
\hline
\texttt{UNIFORM} & 88.5 \\
\hline
\texttt{SMO-MKL} \cite{sun2010multiple} & N/A \\
\hline
\texttt{SwMKL} \cite{kannao2015tv} & 88.5 \\
\hline
\hline
\textbf{Proposed}: (c) + Kernel Dropout & \textbf{90.2}\\
\hline
\end{tabular}
\label{table:restable}
\end{table}

First, we observe that all $3$ kernels achieve accuracies in a similar range, while the fusion of them provides a significant improvement. In the best case, the improvement is over $10\%$ compared to the best of single kernel. Second, we compare each of the proposed training strategies to the standard deep architecture feature fusion (by simple concatenation of learned representations). Dense embedding gives around $4\%$ improvement. This demonstrates the necessity of this preprocessing stage when treating kernel values as embeddings. The inclusion of composition kernels and kernel dropout regularization each provides further improvements. Although the improvement is not tremendous in this case, it is significant. We argue that each step is beneficial and expect much more usefulness of them in more complex problems when a large number of descriptors and kernels are needed. From the visualization of confusion matrix in Figure \ref{fig:confusionMat} we can see the overall classification model is highly effective to this problem and most of the confusion happens only between very related activities (e.g. elevator up versus elevator down).

We compare our approach to MKL methods including combination with uniform weights (denoted as \texttt{UNIFORM}), a popular MKL algorithm \texttt{SMO-MKL} \cite{sun2010multiple} and a recent LMKL approach \texttt{SwMKL} \cite{kannao2015tv}. \texttt{UNIFORM} provides a decent performance and this justifies our utilization of the composition kernels. \texttt{SMO-MKL} applies to binary classification natively and as pointed out by \cite{zien2007multiclass}, the extension to multi-class classification is not trivial. We find this to be true for many existing MKL formulations. \texttt{SwMKL} relies on a regression method to learn the gating function which characterizes the discriminative capabilities of kernels on local data regions. However, in our case each base kernel classifies training data fairly well, causing a highly imbalanced regression problem. This prevents the Support Vector Regressor from obtaining a meaningful gating function, thereby resulting in a performance similar to that of \texttt{UNIFORM} fusion. The proposed approach achieves the best performance and more importantly, provides a reliable way to fuse a large number of kernels in a multi-class setting using powerful numerical and computational backends that are available for generic neural networks. The architecture is also general so that more advanced techniques can be easily incorporated at certain stages. 
ertain stages. 